# Predicting the outcome of team movements; Player time series analysis using fuzzy and deep methods for representation learning

Omid Shokrollahi [1], Bahman Rouhani [2], Amin Nobakhti [3]

[1] *Amirkabir University, Computer engineering department ,Tehran, Iran*
omidoos@gmail.com

[2] *Amirkabir University, Computer engineering department Tehran, Iran*
rouhani74@gmail.com

[3] *Sharif University of Technology, Electrical Engineering department Tehran, Iran*

Telefax: +98 21 6616 4393,

nobakhti@sharif.ir

## Abstract:

We extract and use player position time-series data, tagged along with the action types, to build a competent model for representing team tactics behavioral patterns and use this representation to predict the outcome of arbitrary movements. We provide a framework for the useful encoding of short tactics and space occupations in a more extended sequence of movements or tactical plans. We investigate game segments during a match in which the team in possession of the ball regularly attempts to reach a position where they can take a shot at goal for a single game. A carefully designed and efficient kernel is employed using a triangular fuzzy membership function to create multiple time series for players' potential of presence at different court regions. Unsupervised learning is then used for time series using triplet loss and deep neural networks with exponentially dilated causal convolutions for the derived multivariate time series. This works key contribution lies in its approach to model how short scenes contribute to other longer ones and how players occupies and creates new spaces in-game court. We discuss the effectiveness of the proposed approach for prediction and recognition tasks on the professional basketball SportVU dataset for the 2015-16 half-season. The proposed system demonstrates descent functionality even with relatively small data.

# 1. Introduction:

Extracting knowledge out of time series and spatio-temporal data for activity recognition, analytics and modeling has been a machine learning challenge [29, 7,32]. New research lines began in scouting with the introduction of players' spatiotemporal locations in sports games analytics [22,13]. Efforts to analyze and model different aspects of team coordinated movements, their tactical behavior, has precedence for more than a decade. Team behaviors are complex and adaptive systems. The dynamic spatial arrangements of players in different matches are known to characterize a team's tactical plan.

This paper goes beyond simply relying on old box score statistical analysis and making inferences by extracting knowledge from players' trajectories. Team tactics and players' spatial formations appear in ball and players tracking paths (or trajectories). It is a rational assumption that having position and velocity information about a state, a relatively small duration scene (referred to as micro-events) for some observed data, suffices to estimate the possibility of their transformations and evolutions into or from other states or forms of micro-events scenes with other velocity and position values.

We investigate game segments during a match that the team in possession of the ball regularly attempts to reach a position where they can take a shot at goal and study different tactics to achieve such a position. For example, teams can vary the intensity of an attack by pushing forward, moving laterally, making long passes, or retreating and regrouping. They might happen at any distance concerning the basket (goal). If there are numerous segments and the study segments are long enough, we are likely to sample a tactical patterns diversity. Alternating spatial formations eventually terminate with an event action.

Knowing precisely where each player was throughout a game spurred exciting questions such as: Can we understand how players move with and without the ball to shape the tactical plans? Are we able to simulate player movements and how they affect other players' arrangements? This work's key innovation for modeling tactical plans lies in its approach to model how players' movements influence each other, how short scenes contribute to other scenes and how players occupy and create new spaces in court during small fractions of time. We provide a framework for the useful encoding of short tactics and space occupations in a more extended sequence of movements or tactical plans.

Sports teams will apply new tactics to improve their performance and computational methods to represent, detect, and analyse tactics. This have been the subject of several recent research efforts in soccer and basketball [14, 28, 3]. Similar data-driven concepts are studied in other prominent works such as [5, 11].

The main question is, how can we use machines to study this phenomenon during any offensive possession [19]. Since the advent of player tracking data, a series of research work has concentrated on applying statistical and machine learning tools to introduce recognition and predictive models and measures [37]. Player tracking data have also been extensively studied by researchers in sports analysis [4,6]. However, there are few works on learning new representations from player trajectories for knowledge extraction, which could be useful in many potential applications such as pattern recognition, prediction, clustering, and retrieval.

Fewell considered how team dynamics could be captured as functions by considering games as networks [9]. By defining players as nodes and ball movements as links, they analyzed the network statistics such as degree and flow centrality across positions and teams. It captures the team dynamic through the flow between the ball handlers but not other coincidental teammates' moving patterns.

Many studies of team play also focus on the importance of spacing and spatial context. Metulini seeks to identify spatial patterns that improve team performance on both the court's offensive and defensive ends [25]. The authors use a two-state Hidden Markov Model to model changes in the convex hull's surface area formed by the five players on the court. The model describes how changes in the surface area are tied to team performance, on-court lineups, and strategy. The interplay between offensive actions, or space creation dynamics (SCDs), and defensive actions, or space protection dynamics (SPDs) are studies by [31,18].

In [36] the authors provide a step-by-step guide for leveraging a data feed from SportVU for one NBA game into visualized components that can model any player's movement on offense. Additionally, Miller proposes the idea of applying a language topic modeling for basketball strategies [26]. His team uses deep learning to represent players' trajectories, but the results do not seem to be promising.

Other scholars carried out research to build mathematical predictive models. For example, Cevrone's models for estimating instantaneous possession value rely on an underlying player movement model [5]. They introduced the EPV concept (the expected number of points obtained by the end of a possession) through stochastic process methodology for the evolution of a basketball possession. They model this process at multiple levels of resolution, differentiating between continuous, infinitesimal movements of players and discrete events such as shot attempts and turnovers.

Motivated by recent the advances in the deep learning, some researches considered providing the trace of trajectories as input to several different neural networks. They consider this a simple yet effective way to encode the spatiotemporal pattern of player motion, decisions, actions, and team behaviors. Kempe et al applied the merge self-organizing map (MSOM) to this data and compared it with dynamic controlled neural networks to detect tactical patterns [17]. Various deep neural nets have been employed for modeling purposes in recent years. Wang [34] used LSTM networks for classifying offensive movements, where as in [15] a Generative adversarial networks is used to predict the future trajectory path. As a simple yet effective way of modeling patterns, [8] applied auto-encoders to generate a representation, called a trajectory embedding, of how individual players move on offense.

Mehrasa et al propose a generic deep learning model, using one-dimensional convolutions for learning discriminative feature representations that learn powerful representations from player trajectories [24]. In [8] a technique a proposed using conditional variational auto-encoders, which learns a model that also "personalizes" prediction to individual agent behavior within a group representation.

Other have worked on the technical problem of improving time series models for learning representation. Hyvarinen and Morioka [16] learn representations on evenly sized subdivisions of time series by learning to discriminate between those subdivisions from these representations. Lei and colleagues [20] exploit an unsupervised method in which the distances between learned representations mimic a standard distance (Dynamic Time Warping, DTW) between time series. Malhotra et al [23] design an encoder as a recurrent neural network, jointly trained with a decoder as a sequence-to-sequence model to reconstruct the input time series from its learned representation. Finally, [35] compute feature embedding generated in approximating a carefully designed and efficient kernel.

Most previous deep learning researches build their model, taking input trajectories data as a single picture, while some feed raw players' time series into a model. In this work, we propose a combination of deep and fuzzy methods on players tracking time series data. It appears that this hybrid model outperforms the algorithm by [24] for representation learning.

# 2. Methodology

## 2.1. Datasets and Data mining

The STATS SportVU dataset for the 2015-2016 half-season, consists of players' positions and the ball in 2D world coordinates captured by a six-camera system at a frame rate of 25Hz. Each frame does not have complete annotations of the events happening in this frame, such as fouls, possession, shot, pass, and rebound. To extract independent annotated data records to be used as game segments, this dataset's information is aligned in time with the "Play by Play" dataset recording the events occurring during the match. Using the "Play by Play" dataset, we have the exact time for event action types. We extract the time entries within these intervals as the independent event records. A full description of this raw data and different data mining phases to process into a time series table is found in [36].

The next goal is to extract independent records or data segments out of the continuous players' path. Separate, distinct segments should start at one event action and terminate at the following one. As these events vary in their size and final event action type, only those longer than 1 second are selected for preprocessing as legitimate events. One approach to extract equally sized records is to cut a specific length tail of these time frames. Another is to slide a fixed-sized window on all events. Having these frames with equal length records is not necessary to train our time series neural network model, but doing so lays a more standard foundation for extracting independent records or data segments for the experiment.

At this stage, multiples of short-lasting fixed-size micro-events are extracted from each event as we slide a 1- second duration time window on the sequence. Event time series are divided into sub-events with a stride of 0.2 seconds. Accordingly, events may have various durations, but sub-events will necessarily have a length of 1 second. These parameters can change for different setups, and our encoder architecture has no restriction for this issue. All such derived sub-events from an event will be assigned to the same label, which is the action that occurs at the end of the main event.

The set of Labels (classes) of actions in the game data set we consider are:

- Make (26 events)
- Miss: (29 actions)
- Fouls: (11 actions)
- Out of bound + turnover + steal (losing the ball): 17 actions

There are around 8000 sub-events for train and test experiment, extracted from one single game. The Makes and Misses are combined into the Shot class since logically passing the ball from the basket does not appear to be correlated to the patterns of trajectories and our experiment setting.

## 2.2. Applying the "potential of presence" kernel

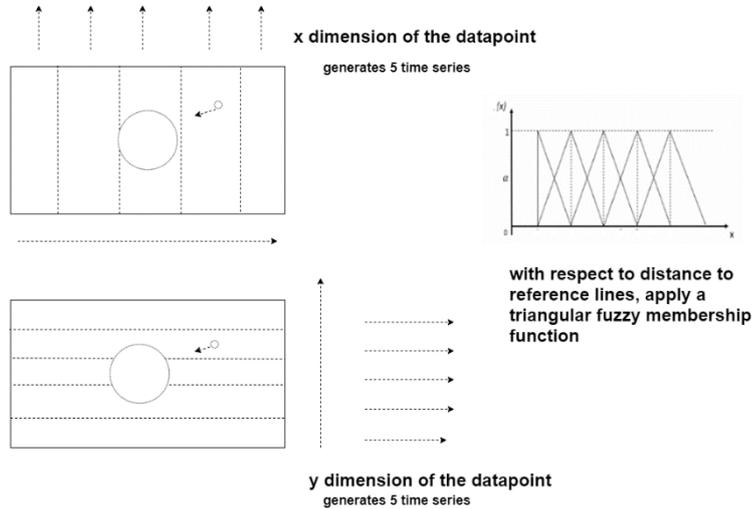

Fig 1 the kernel for high dimensional time series

Prior feeding the players' trajectories as multivariate time series into the network, we extract features that properly represent the essence of team tactical behavior. We use a fuzzy system and fuzzy time series to represent players' positions with respect to different court regions. It assigns a set of belief values to each time series's ranges concerning the player's distance to some predefined lines in the court along with both $x$ and $y$ directions.

Players usually arrange in one, two, or three offensive-defensive layers (along the $x$-direction), and may be distributed in two to five lateral sections (along the $y$-direction). Consequently, we divide the $x$-axis and $y$-axis of the court rectangle into five regions (figure1) and applied the Triangular fuzzy membership function on series to assign fuzzy values for the distance of balls and players for each mentioned line. The triangular membership function is proven to work best for one-dimensional series. As the player's distance or the ball at these regions concerning the basket changes, the belief values time series for some sub-regions rise, and others' values drop or is set to zero.

Time-series are primarily related to the $x$ and $y$ dimension of the ball, five players of the current team, and five players of the opponent for each event (22 dimensions). When membership functions are applied for these values, each dimension creates five new dimensions (in sum 110 dimensions). The generated time series of the original time series somehow represents the potential of being present (or occupying) at each region. The belief value serves as a potential function for the presence of players. We will use these presence-potential time series to improve the distinction between different tactical patterns.

The 110 time-series information regarding a short sequence of snapshots of trajectories is extracted and then presented using an encoder described in the next section. Thus with the multivariate time series representation model, every sub-event has a corresponding vector, demonstrated by a point in a high dimensional space. This embedded space has some interesting characteristics. One is the Representations metric space. Besides being suitable for classification purposes, the learned representations may also define a meaningful measure of the distance between time series.

## 2.3. Unsupervised scalable representation learning for multivariate time series:

We use an algorithm [10] that outperforms the previously mentioned methods, also tested on many standard datasets for unsupervised representation learning for time series. This model trains an encoder-only (without decoder) architecture to reduce the computational cost.

**Triplet Loss:** It is a loss function for machine learning algorithms where a baseline (anchor) input is compared to a positive (truthy) input and a negative (falsy) input. The distance from the baseline (anchor) input to the positive (truthy) input is minimized, and the distance from the baseline (anchor) input to the negative (falsy) input is maximized. It is widely used in various forms for representation learning in other domains such as NLP [12,33]. It has also been theoretically studied [1] in a fully unsupervised problem, as some existing works assume the existence of class labels or annotations in the training data. This method instead relies on a more natural choice of positive samples, learning similarities using subsampling. The proposed triplet loss uses original time-based sampling strategies to overcome the challenge of learning on unlabeled data. The triplet loss objective function ensures that similar time series obtain similar representations, with no supervision to learn such similarity.

Following the analogy with word2vec, $x^{pos}$ corresponds to a word, $x^{ref}$ to its context, and $x^{neg}$ to a random word. The Objective corresponding to these choices is presented. A deep network $f(\cdot, \theta)$ with parameters $\theta$ and the $\sigma$ is the sigmoid function. This loss pushes the computed representations to distinguish between $x^{ref}$ and $x^{neg}$ and to assimilate $x^{ref}$ and $x^{pos}$.

$$Loss = -\log(\sigma(f(x^{ref},\theta)^T f(x^{pos},\theta))) - $$
$$\sum_{k=1}^{K} \log(\sigma(-f(x^{ref},\theta)^T f(x_k^{neg},\theta))) \quad (1)$$

The corresponding loss then pushes pairs of (context, word—which should be close) and (context, random word-which should be considered unrelated) to be linearly separable. This is called negative sampling.

$x^{ref}$ is a random subseries of a given time series $y_i$. The representation of $x^{ref}$ should be close to the one of any of its subseries $x^{pos}$ (a positive example). If one considers another subseries $x^{neg}$ (a negative example) chosen at random (in a different random time series $y_j$ if several series are available, or in the same time series if it is long enough and not stationary), then its representation should be distant from the one of $x^{ref}$. Several negative examples are chosen here.

Overall, the training procedure consists of traveling through the training dataset for several epochs (possibly using mini-batches), picking tuples $(x^{ref}, x^{pos}, (x_k^{neg})_k)$ at random as detailed in Algorithm 1 of franceschi [10], and performing a minimization step on the corresponding loss for each pair, until training ends. This unsupervised training is scalable as long as the encoder architecture is scalable as well.

The length of the negative examples is chosen at random in Franceschi's Algorithm 1 for the most general case; however, their length can also be the same for all samples and equal to size($x^{pos}$). The latter case is suitable for our experiment where all time-series in the dataset have equal lengths of the sliding window. It is time and memory-efficient both on train and test.

**Algorithm 1:** Choices of $x^{ref}$, $x^{pos}$ and $(x_k^{neg})_{k \in [1,K]}$ for an epoch over the set $(y_i)_{i \in [1,N]}$.

1. **for** $i \in [1, N]$ with $s_i = \text{size}(y_i)$ **do**
2.     pick $s^{pos} = \text{size}(x^{pos})$ in $[1, s_i]$ and $s^{ref} = \text{size}(x^{ref})$ in $[s^{pos}, s_i]$ uniformly at random;
3.     pick $x^{ref}$ uniformly at random among subseries of $y_i$ of length $s^{ref}$;
4.     pick $x^{pos}$ uniformly at random among subseries of $x^{ref}$ of length $s^{pos}$;
5.     pick uniformly at random $i_k \in [1, N]$, then $s_k^{neg} = \text{size}(x_k^{neg})$ in $[1, \text{size}(y_k)]$ and finally $x_k^{neg}$ among subseries of $y_k$ of length $s_k^{neg}$, for $k \in [1, K]$.

Franceschi Algorithm 1- for time series embedding [10]

We use deep neural networks with exponentially dilated causal convolutions to handle time series. The model is mainly based on stacks of dilated causal convolutions (see Figure 2a), which map a sequence to a sequence of the same length, such that the i-th element of the output sequence is computed using only values up until the *i-th* element of the input sequence, for all *i*. It is thus called causal since the output value corresponding to a given time step is not computed using future input value.

**Convolutional networks for time series:** Dilated convolutions, popularized by WaveNet [27], were shown to perform as well as sequence-to-sequence models for time series forecasting [2] using an architecture that inspired this model. These work mainly show that dilated convolutions help to build networks for sequential tasks outperforming recurrent neural networks in terms of both efficiency and prediction performance.

Inspired by [2], we build each layer of our network as a combination of causal convolutions, weight normalizations [30], leaky ReLUs, and residual connections (see Figure 2b). Each of these layers is assigned an exponentially increasing dilation parameter ($2^i$ for the *i*-th layer). The output of this causal network is then given to a global max-pooling layer squeezing the temporal dimension and aggregating all temporal information in a fixed-size vector (as proposed by [33] in a supervised setting with full convolutions). A linear transformation of this vector is then the encoder's output, with a fixed, independent from the input length, size.

Causal convolutions allow alleviating the disadvantage of not using recurrent networks at testing time. Indeed, recurrent networks can be used in an online fashion, thus saving memory and computation time during testing. In our case, causal convolutions organize the computational graph so that, to update its output when an element is added at the end of the input time series, one only has to evaluate the highlighted graph shown in Figure 2a rather than the full graph.

This encoding methodology is both scalable and efficient both at training and inference time; this works for both short and long time series. This fact makes it suitable for experimenting with an arbitrary window size length.

The reader can find a description of the whole process via algorithm 1 in appendix.

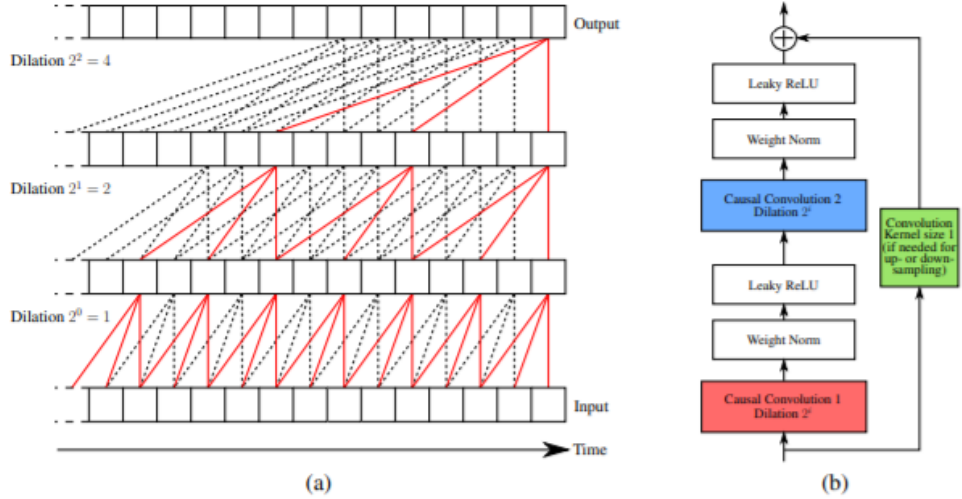

Fig 2 (a) Illustration of three stacked dilated causal convolution. Lines between each sequence represent their computational graph. Red solid lines highlight the dependency graph for the computation of the last value of the output sequence, showing that no future value of the input time series is used to compute it. (b) composition of the i-th layer of the chosen architecture [10]

## 3. Results and Discussion

We evaluate the performance of the proposed model under two main experimental setup. Fr the first experiment, we consider all micro-events as samples of a population. The high accuracy scores prove that it is useful for post-game analysis. We have access to all the appeared data patterns and the objective is to make inferences about the trajectory patterns similar to the observed data. In this setup, we adopt a range of percentages as the random training set and leave the rest as the test data. It is evident that the encoding and classification require separate training procedures. Since the encoding algorithm is unsupervised, and data can be encoded beforehand for an arbitrary portion of the data, we may consider the classification on different sets. Thus this setup can be divided into branching experiments. In the first one, we train the encoder with 80% of the data and then test a range of data portions. The result was slightly better than when we train the encoder with a percentage of the data and train the classifier on the same data. The diagram shows the case.

In both experiments, we consider all micro-events samples of a homogenous population of different patterns. A wide range of patterns may appear in the selected selection; they may belong to events that had distinct inherent tactical patterns. For 80% of the data, we get more than 94% accuracy. The confusion matrix also testifies that all classes of data are recognized with more than 90% accuracy. As we reduce the training set's size to 5%, the accuracy drops to 82%, but the confusion matrix shows strong classifications for all action classes. To explain this pattern, we may deduce that the algorithm observes fragmented segments of a wide range of tactical patterns. It can guess the encoding for the unseen parts of the trajectories using the inferred encodings for micro-events. Some errors are acceptable since the initial part of team movements (in all classes) may be random with no tactical plan. Overall the results appeared to be stable with adopting different random data to study.

We note that micro-events overlap with their following and preceding four micro-events in the event time series. This fact promotes their correlation and therefore encoding strength. This restriction is helpful for the encoding mechanism; it forces the game's connected micro-events to lie in near regions in the representation space. However, based on the relatively good results when using 5% training set (in which the micro-events are not expected to overlap) illustrate that the test cases do not necessarily need to overlap with the training data.

We also perform another experiment where we used the model to predict micro-events belonging to events with an unobserved tactical plan. The result was not as accurate, which is expected since the algorithm can only make sound judgments for test cases similar to the observed tactical patterns. Theoretically, patterns may embody the arbitrary positioning and composition of players and the ball; consequently, the trajectories may differ significantly. Therefore our encoder is not expected to be helpful for the case. For this setup, it is not a fair expectation from our model to make reasonable judgments. Bornn's team [24] report a similar event classification via representation learning. We did not have access to their dataset, where labeled events were collected from many games, so these two experiments are not entirely comparable.

| Confusion matrix: Network 80%,classifier 80% | Class 1 | Class 2 | Class3 |
|---|---|---|---|
| Class 1 | 1162 | 12 | 8 |
| Class 2 | 26 | 150 | 5 |
| Class 3 | 5 | 1 | 168 |

Table 1- confusion matrix for the first experiment setup 80% data for training network,same 80% for classifier

| Confusion matrix: Network 80%,classifier 5% | Class 1 | Class 2 | Class 3 |
|---|---|---|---|
| Class 1 | 4936 | 274 | 401 |
| Class 2 | 452 | 322 | 84 |
| Class 3 | 488 | 87 | 508 |

Table 2- confusion matrix for the first experiment setup, 80%data for training network,5% for classifier

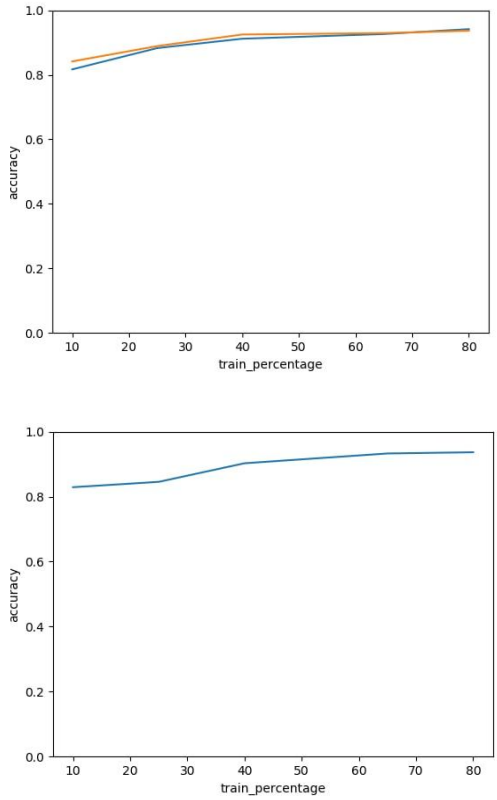

Fig 3 plotting accuracy for two different experiment setups. (a)80% of data for training and different percentage for classifier (up). (b)accuracy for a training percentage and the same data for classifier (down).

| % of trained data | Accuracy of Network, train(80%) classifier train(% of entire dataset) | Accuracy of Network, train (% of entire data set), classifier train (the same data) |
|---|---|---|
| 80 | **0.941542288557213** | 0.936567164179104 |
| 65 | 0.926386913229018 | 0.932788051209103 |
| 40 | 0.911825726141078 | 0.902697095435684 |
| 25 | 0.882821576763485 | 0.845477178423236 |
| 10 | 0.816848803430626 | 0.829021994743394 |
| 5 | 0.7635063559322034 | 0.801906779661017 |

Table 3- accuracy for two different experiment setup

# 4. Conclusion:

We introduced a kernel and an encoder to represent the players' coordinated trajectories to model the tactical patterns of team players in basketball. Our approach was a hybrid of fuzzy and deep convolutional neural networks for the multivariate time series data, which maps the players and the ball trajectories to a high dimensional space. The data is extracted from the sportVU player tracking dataset, which is synced with the "Play-by-play dataset" that recorded the events that occur during the match. After combining these two datasets, we used our model to predict the eventual action that happens at the end of small segments of game events.

We expected our algorithm to make inferences about relatively similar micro-events to those observed during the training. The results show that we had above 90% accuracy for each class. It shows that our algorithm can be used for other test cases where the data is slightly different compared to the training data. Inference makes sense in this case, and it is one of the advantages of our model that it exhibits good results with relatively small data belonging to a single game. Accordingly, even though we have built a robust kernel and encoder, it does not seem useful to predict the outcome of drastically different unseen tactical patterns. A rule based algorithm may be more useful for that case. The prediction may also be explored in a semi-supervised manner, which is proposed as future works. It is also possible to take advantage of the meaning of distance between the representation of micro-tactics for other analytical purposes. For example, one can explore the similarity between different tactical patterns, cluster them, or explore the path it takes to move from one tactical state to another.

## Future work:

Our micro-events time series are rarely or sparsely labeled. Not all the micro-events should be labeled similar to the action at the end of the micro event because technically, they may contribute to alternative outcomes and have a vague label. We have an unsupervised representation learning for the extracted time series. We should use a different classifier for a semi-supervised setting that will be studied in our future experiments.

If we create an edge between enough-near points, communities represent a class of micro-tactics. (as future work, we can study the transition between micro-tactics using link prediction and community detection)

## Appendix:

Applying Triangular fuzzy membership function:

As it was discussed in the methodology, we apply a fuzzy membership function on the x, y time series of the players and the ball. The process is described as follows. The x, y dimension of the court is (94 , 50)

*max(min((x - a) / (b - a), (c - x) / (c - b)), 0)*

a,b,c are the three values to represent the triangular shape of the membership function.

| X-dimension Table a | Param_a | Param_b | Param_c | Y-dimension Table b | Param_a | Param_b | Param_c |
|---|---|---|---|---|---|---|---|
| A | 98 | 94 | 72 | M | -2 | 0 | 17 |
| B | 94 | 72 | 47 | N | 0 | 17 | 33 |
| C | 72 | 47 | 22 | O | 17 | 33 | 50 |
| D | 47 | 22 | 0 | P | 33 | 50 | 51 |
| E | 22 | 0 | -4 | Q | 50 | 51 | 60 |

Table 4- parameters for our triangular fuzzy membership function

| **Algorithm 1: workflow** |
|---|
| **1-** Extract action time for makes, miss, outofbound/turnover/steal and fouls from play by play dataset |
| **2-** Filter the time series, according to time to create the events (start= action time, end=next action) |
| **3-** for filtered data, label=next action. |

**4-** Move a Sliding window with length =*l* sec and stride =*s* sec on every event, store the generated frames as micro- events,

**5-** Label each micro event with the next action

**6-** Discard the events with length less than *l* second

**7-for** *x* in the collection of *X* time series of micro-events:

apply *z* fuzzy membership function, parameters given in the appendix table(a), store as $X_{series}$

**8-for** *y* in the collection of *Y* time series of micro-events:

apply *w* fuzzy membership function, parameters given in the appendix table(b), store as $Y_{series}$

**9-for** $Z_{train}$: *m* percentage of the whole data:

randomly select *m* percentage of the entire micro event as $Z_{train}$, and 100-*m* as $Z_{test}$

**10-** feed $z * (Count_{players} + 1)$ dimensions of $X_{series}$ and $w * (Count_{players} + 1)$ dimension of the $Y_{series}$ of $Z_{train}$ together to the neural net (here **1** in the formula represents the ball) to generate the embedding vector for $Z_{train}$ and $Z_{test}$

**11-for** classification train size *v* as $C_{train}$, keep the rest as $C_{test}$:

use the embeddings $Z_{train}$ & $Z_{test}$ and corresponding labels for $Z_{train}$, predict labels of $Z_{test}$ with SVM or KNN classifier

**12-** Separately use $Z_{train}$ and $Z_{test}$ and labels of $C_{train}$ to predict the labels for $C_{test}$ with SVM or KNN